\documentclass[letterpaper, 10 pt, conference]{ieeeconf}  %
\IEEEoverridecommandlockouts                              %
\usepackage{graphicx} 
\usepackage{algorithm}
\usepackage{algorithmic}
\usepackage{lipsum} 
\usepackage{flushend}
\usepackage{float}
\usepackage{booktabs}
\usepackage{xcolor}
\usepackage{hyperref}

\usepackage{amsmath}
\usepackage{amsmath,amssymb,amsfonts}
\usepackage{algorithmic}

\usepackage{textcomp}
\usepackage{pdflscape}
\usepackage[table]{xcolor}
\usepackage{subfigure}
\usepackage{pifont}
\usepackage{placeins} 
\usepackage{bm}

\begin{document}

\title{AIMC-Spec: A Benchmark Dataset for Automatic Intrapulse Modulation Classification under Variable Noise Conditions}

\author{Sebastian L. Cocks, Salvador Dreo, Brian Ng, and Feras Dayoub
\thanks{S. L. Cocks, B. Ng, and F. Dayoub are with Adelaide University, Adelaide, SA 5005 AUS (e-mail: sebastian.cocks@adelaide.edu.au).}%
\thanks{The first version of this work was published in IEEE Access DOI: 10.1109/ACCESS.2025.3645091}}


\maketitle

\begin{abstract}

A lack of standardized datasets has long hindered progress in automatic intrapulse modulation classification (AIMC)—a critical task in radar signal analysis for electronic support systems, particularly under noisy or degraded conditions. AIMC seeks to identify the modulation type embedded within a single radar pulse from its complex in-phase and quadrature (I/Q) representation, enabling automated interpretation of intrapulse structure. This paper introduces AIMC-Spec, a comprehensive synthetic dataset for spectrogram-based image classification, encompassing 30 modulation types across 5 signal-to-noise ratio (SNR) levels. To benchmark AIMC-Spec, five representative deep learning algorithms—ranging from lightweight CNNs and denoising architectures to transformer-based networks—were re-implemented and evaluated under a unified input format. The results reveal significant performance variation, with frequency-modulated (FM) signals classified more reliably than phase-modulated (PM) types, particularly at low SNRs. A focused FM-only test further highlights how modulation type and network architecture influence classifier robustness. AIMC-Spec establishes a reproducible baseline and provides a foundation for future research and standardization in the AIMC domain.

\end{abstract}


\vspace{10pt}

\maketitle

\section{Introduction}
\PARstart{E}{fficient} and accurate analysis of electromagnetic signals is critical to a wide range of modern applications, including spectrum monitoring, cognitive electronic warfare, and other advanced signal processing systems. These tasks often involve detecting, classifying, and interpreting signals in complex and noisy environments. Typically, raw electromagnetic emissions are captured as I/Q data, which can then be processed into Pulse Descriptor Words, encoding features such as frequency, pulse width, and time of arrival, or transformed into time–frequency spectrograms for visual and algorithmic analysis. These representations form the basis of automated techniques for identifying signal characteristics and modulation types, enabling reliable performance in dynamic spectral conditions \cite{RN136, RN153, RN160, RN163, RN156}.

One critical classification task in this context is AIMC--the process of identifying the modulation scheme contained within a single radar pulse from its complex baseband representation. Formally, given a discrete I/Q sequence

\[x[n]=I[n]+jQ[n], \: n=1,2,...,N\]

where each pulse $p_i$ within $x[n]$ possesses an underlying modulation type $m_i$, the objective of AIMC is to estimate the corresponding labels $\widehat{m}_i$ such that $\widehat{m}_i{\approx}m_i$ for all pulses $p_i{\in}x[n]$. In practical terms, this involves transforming the I/Q data into a suitable time-frequency representation and applying a classification model to determine the modulation type present. 

AIMC is a specialized sub-problem of radar modulation classification focused on intrapulse characteristics rather than interpulse or emitter-level behavior. Its correct formulation is essential for building robust radar signal analysis and electronic intelligence systems.

Prior to the emergence of deep learning, AIMC methodologies primarily followed a two-step process: feature extraction and classification. Feature extraction involved techniques such as time-frequency transformations (e.g., STFT, WVD, CWD) and statistical analyses to isolate and represent key signal characteristics \cite{Cai2022RadarIS, Akyon2018ClassificationOI}. These features were then input into classifiers like SVMs, k-NN, or probabilistic models \cite{Yuan2021IntraPulseMC}. While these traditional methods provided foundational insights, they often required extensive domain knowledge and struggled with low SNR conditions, limiting their robustness and scalability. 

While there has been progress in applying deep learning and image-based classification to AIMC, the field lacks a standardized, publicly available benchmark dataset. Most existing studies rely on private or internally generated data with limited transparency, making fair comparisons and reproducibility difficult.

The AIMC-Spec dataset was previously introduced as a synthetic benchmark for reproducible evaluation of intrapulse modulation classification algorithms \cite{REF_AIMC_SPEC_V1}. While the initial release provided structured coverage of a broad range of modulation types and SNR conditions, its practical use revealed several limitations. In particular, the original storage format resulted in a very large dataset footprint, exceeding 500\,GB for the full release, which hindered accessibility and rapid experimentation. This work presents an updated version of AIMC-Spec that addresses these limitations through a redesigned data storage pipeline, reducing the dataset size to under 30\,GB while preserving full signal fidelity. In addition, the modulation set has been refined and the dataset generation process streamlined to improve usability, consistency, and reproducibility for AIMC benchmarking.

This paper presents the updated AIMC-Spec dataset as a synthetic benchmark for spectrogram-based AIMC using image classification methods. It includes a diverse collection of modulation types, including both radar and communication waveforms, and covers a wide range of SNR levels to support consistent and structured evaluation. The dataset is fully synthetic, allowing complete control over pulse parameters and noise levels while ensuring consistent ground-truth labels. This design isolates algorithmic performance from hardware-specific or environmental variables. To demonstrate the dataset’s practical utility, five representative deep learning algorithms from recent AIMC literature were selected, re-implemented, and benchmarked under varied signal conditions, enabling comparative analysis across architectures, modulation types, and noise levels. A focus of this work is on how different deep learning architectures respond to a common spectrogram representation of radar signals, rather than on comparing transform or preprocessing strategies. This framing ensures that observed performance differences reflect architectural design rather than input representation bias. 

\subsection{Contributions}
\begin{itemize}
    \item This work addresses a major gap in the AIMC field by introducing a publicly available synthetic dataset, AIMC-Spec, designed specifically for benchmarking image-based classification models. It includes 30 distinct intrapulse modulations and standardized signal definitions to support consistent comparison across future studies.
    
    \item Benchmarking results are provided using five diverse AIMC approaches, covering varied modeling strategies and preprocessing pipelines. These evaluations span 5 SNR levels, with an additional focused assessment on FM-only classification to isolate model performance under constrained conditions.
    
    \item All data and code are openly accessible for reproducibility and future research: \url{https://www.kaggle.com/sebastiancocks/datasets?query=aimc-spec} and \url{https://github.com/seb-cocks/AIMC-image-classification}
\end{itemize}

\begin{figure*}[t]
    \centering
    \includegraphics[width=\textwidth]{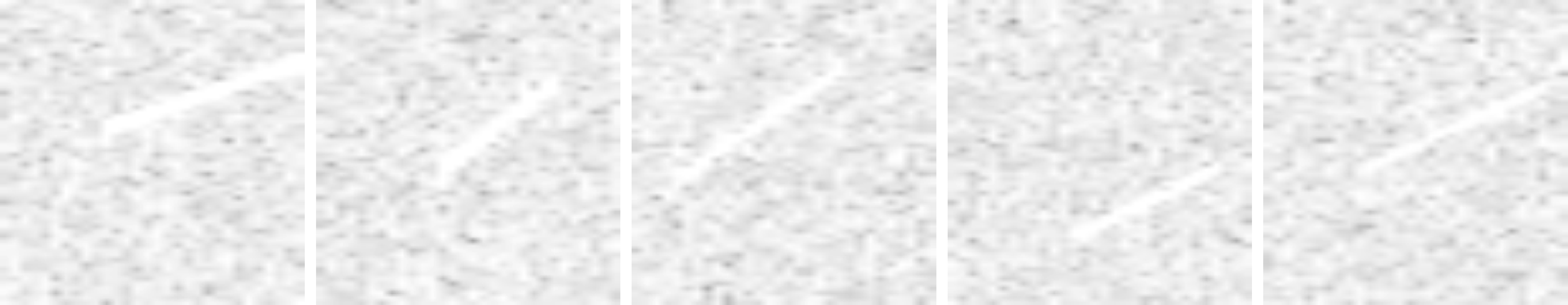}
    \caption{Spectrograms of an LFM-up signal across the SNR levels used in AIMC-Spec, ranging from +6\,dB (left) to –6\,dB (right) in 3\,dB steps. The progression illustrates the gradual degradation of the characteristic linear frequency sweep as the noise level increases.} \label{fig:snr-samples}
\end{figure*}

\section{Related Work}

Research in AIMC has increasingly turned to deep learning, leveraging image-based representations of radar signals for robust classification under noisy and complex conditions. A broad range of models have been proposed, often distinguished by their architectural design, spectrogram generation techniques, and strategies for handling noise. While several datasets have been developed for radar or communication signal analysis, the AIMC field continues to lack a standardized, task-specific dataset to support fair and reproducible benchmarking.

Notable datasets such as RadarCommDataset \cite{Jagannath2021DatasetFM}, RadChar \cite{Huang2023MultiTaskLF}, and RadioML \cite{oshea2017} include, but are not limited to, intrapulse modulations like linear frequency modulated (LFM), Barker and binary phase shift-keying (BPSK). However, they are not tailored to the AIMC problem. These resources often provide limited modulation coverage, do not consistently define signal structures, or focus more broadly on mixed communication and radar domains. As a result, despite their value, they fall short of supporting the reproducibility and comparability needed for rigorous AIMC evaluation. The development of a dedicated, well-documented AIMC dataset remains a necessary step to advance the field.

Several studies have focused on refining input quality through spectrogram preprocessing. Wang et al.~\cite{RN136}, proposed a lightweight CNN model using Short-Time Fourier Transform (STFT) spectrograms enhanced by multi-stage zero-mean scaling and noise reduction, achieving 96.7\% accuracy at an SNR of –6\,dB across 6 modulation types. He et al.~\cite{RN161}, used Continuous Wavelet Transform (CWT) to produce high-resolution time-frequency images, reaching 90\% accuracy at –11\,dB with a compact CNN evaluated on 12 modulation types. The work of Kong et al.~\cite{RN151} introduced the Fourier synchrosqueezing transform (FSST) in place of Choi-Williams Distribution (CWD), applying complex-valued spectrograms for CNN classification and attaining 98.4\% accuracy at –10\,dB on 8 modulation classes. The LPI-Net model, developed by Huynh-The et al.~\cite{RN162}, utilized CWD spectrograms in a modular CNN built from asymmetric convolutional paths and skip connections, achieving 98.6\% accuracy at 0\,dB across 13 modulation types. Across these studies, all models were trained on custom datasets that were not made publicly available. Reported accuracies ranged from 90\% to 98.6\%, typically evaluated at SNR levels between –11\,dB and 0\,dB, with class counts ranging from 6 to 13.

Other studies incorporated denoising mechanisms into the learning pipeline. The approach by Qu et al.~\cite{RN153} combined a Convolutional Denoising Autoencoder (CDAE) with an Inception-based CNN to denoise Cohen’s Class Distribution (CTFD) inputs, achieving over 95\% accuracy at –9\,dB across 12 modulation types. A DenseNet classifier in conjunction with a CDAE was proposed by Yang et al.~\cite{RN148}, reaching 99.6\% accuracy at –6\,dB and 92\% at –10\,dB on 6 modulation classes using STFT spectrograms. The LDC-Unet model by Jiang et al.~\cite{RN163} integrated a U-Net feature extractor with local dense connections and a fine-tuned VGG-19 classifier trained with a custom SSR-loss function, achieving 91.17\% accuracy at –10\,dB across 12 modulation types. All three studies used custom datasets that were not made publicly available. Reported accuracies ranged from 91.17\% to 99.6\%, with evaluations commonly conducted at SNR levels between –10\,dB and –6\,dB, covering 6 to 12 modulation classes.

Transformer components and attention mechanisms have also emerged in recent architectures. RM-Net of Tang et al.~\cite{RN156} combined ResNet blocks with multi-head self-attention and SCR-loss on CWD spectrograms, achieving 95.5\% accuracy at –10\,dB across 12 modulation types. The method by Shen et al.~\cite{RN157} employed a parallel residual CNN trained on CWD spectrograms under multi-path fading environments, reporting 81\% accuracy at –10\,dB on 12 classes. Both studies evaluated performance on private datasets that were not publicly released. Despite the architectural differences, they shared common evaluation conditions—namely, 12 modulation types and low-SNR scenarios centered around –10\,dB—highlighting their focus on robustness under challenging channel conditions.

Additional strategies include multi-label classification and joint learning. Qu et al.~\cite{RN160} used Deep Q-Learning with custom CTFD kernels, achieving 94.5\% accuracy at –6\,dB across 8 modulation types. A DnCNN-ResNet pipeline for aliasing-resilient multi-label classification was presented by Hong-hai et al.~\cite{RN152}, reporting 97\% accuracy at 0\,dB on 5 classes. The approach of Li et al.~\cite{RN159} employed Kernel Collaborative Representation and Discriminative Projection alongside a fine-tuned CNN, reaching 97.58\% at –6\,dB across 12 modulation types. A semi-hard triplet loss and SER block-based method was introduced through Qu et al.~\cite{RN149}, achieving 88\% accuracy at –10\,dB for unknown radar signal recognition across 14 modulation types. These models were all evaluated using internal datasets that were not publicly shared. Accuracy levels ranged from 88\% to 97.58\%, with class counts varying widely—from 5 to 14—depending on whether the task targeted known, unknown, or overlapping signal types.

Recent work by Bhatti et al.~\cite{Bhatti2024TransformerbasedMF} explored transformer-based architectures for recognizing phase-coded radar waveforms under extremely low SNR conditions. Their study proposed three AMRS models—ViT, Vicinity ViT (VViT), and a CNN baseline—trained on phase spectrograms derived from STFT. Focusing on six types of phase-coded signals—Barker, Frank, and Polyphase 1--4 (P1--P4)—they demonstrated robust recognition at SNRs as low as --16\,dB. The VViT model achieved a recognition accuracy of 93\% at –16\,dB, outperforming ViT (92.7\%) and CNN (89\%). By leveraging phase information rather than magnitude, this work highlights the potential of transformer-based models in ultra-low SNR AIMC scenarios.

From this broad survey, five representative approaches \cite{RN136,RN162,RN153,RN163, Bhatti2024TransformerbasedMF} were selected for benchmarking. These encompass lightweight CNNs, denoising autoencoders, modular convolutional pipelines, fine-tuned pretrained backbones, and transformer-based networks, capturing the methodological diversity of AIMC research and enabling comparative evaluation on a consistent dataset. 

Despite these promising developments, the lack of common datasets and inconsistencies in reported signal definitions, SNR conditions, and generation methods continue to impede progress. Tables~\ref{tab:mod_presence_fm}, \ref{tab:mod_presence_pm} and \ref{tab:mod_presence_hybrid} illustrates the limited overlap in modulation types across studies, emphasizing the need for standardized datasets to ensure fair comparison and reliable benchmarking. 

\subsection{Case for New Dataset}

Despite the range of deep learning approaches applied to AIMC, nearly all rely on custom datasets that are not publicly available. These datasets vary in modulation class count, SNR range, and preprocessing methods, making fair comparison and reproducibility difficult. Reported results span from 5 to 14 classes and SNRs from 0\,dB to –11\,dB, with no consistent evaluation standard across studies.

While existing resources like RadarCommDataset \cite{Jagannath2021DatasetFM}, RadChar \cite{Huang2023MultiTaskLF}, and RadioML \cite{oshea2017} offer intrapulse modulations, they are not designed for AIMC. Their broader focus and inconsistent definitions limit their use for benchmarking intrapulse classification tasks. To address these gaps, this paper introduces AIMC-Spec: a dedicated dataset for reproducible, standardized evaluation of AIMC methods.

\section{Data Set Properties}
AIMC-Spec is divided into 5 SNR levels, each containing 30 intrapulse modulations with 1{,}000 testing signals and 100{,}000 training per modulation. Signals were generated via a Python script that automatically selected parameters relevant to each modulation type. This section outlines the key dataset attributes, including signal configurations, noise conditions, modulation types, and file organization.

\subsection{Pulse and Modulation Parameter Selection}
Each signal sample contains only the pulse itself, with no pure noise-only segments stored. The pulse width (PW) is uniformly distributed between 0.5 $\mu{s}$ and 18 $\mu{s}$, with a resolution of $1\;ns$. Signals are sampled at a frequency of 50 MHz. The carrier (RF) frequency is uniformly distributed within a 50 MHz bandwidth centered at 9500 MHz, with a valid step size of 0.001 MHz (1 kHz resolution). The dataset is constructed with a uniform distribution across all valid parameter combinations. To enforce diversity and prevent over-representation of specific parameter pairs, each modulation type allows at most three repeated instances of any particular PW–RF combination.

\begin{table*}[htbp]
\caption{FM class usage across referenced studies. Each row represents a modulation type and each column a cited paper. Filled cells indicate the modulation was included in that study; empty cells indicate it was not.}
\label{tab:mod_presence_fm}
\centering
\footnotesize
\resizebox{\textwidth}{!}{%
\begin{tabular}{|l|l|c|c|c|c|c|c|c|c|c|c|c|c|c|c|c|c|c|c|c|c|c|l|}
\hline
 &  AIMC-Spec&\cite{RN161} & \cite{RN162} & \cite{RN152} & \cite{RN163} & \cite{RN151} & \cite{RN159} & \cite{RN160} & \cite{RN153} & \cite{RN149} & \cite{RN157} & \cite{RN156} & \cite{RN136} & \cite{RN148} & \cite{Hoang2019AutomaticRO} & \cite{Qu2018RadarSI} & \cite{Wang2017AutomaticRW} & \cite{Kong2018AutomaticLR} & \cite{Zhang2017ConvolutionalNN} & \cite{Tang2025AutomaticRO} & \cite{Chen2024LPIRS} & \cite{Luo2024RadarWR}  &\cite{Bhatti2024TransformerbasedMF}\\
\hline
UNMOD &   \cellcolor{black}&& \cellcolor{black} &  &  &  &  &  &  &  &  &  & \cellcolor{black} &  &  &  &  &  &  &  &  & \cellcolor{black}  &\\
\hline
LFM &  \cellcolor{black}&\cellcolor{black} & \cellcolor{black} &  & \cellcolor{black} & \cellcolor{black} & \cellcolor{black} & \cellcolor{black} & \cellcolor{black} & \cellcolor{black} & \cellcolor{black} & \cellcolor{black} & \cellcolor{black} & \cellcolor{black} & \cellcolor{black} & \cellcolor{black} & \cellcolor{black} & \cellcolor{black} & \cellcolor{black} & \cellcolor{black} & \cellcolor{black} & \cellcolor{black}  &\\
\hline
NLFM &  \cellcolor{black}&\cellcolor{black} &  &  &  &  &  &  &  &  &  &  & \cellcolor{black} & \cellcolor{black} &  &  & \cellcolor{black} &  &  &  & \cellcolor{black} & \cellcolor{black}  &\\
\hline
SFM &   \cellcolor{black}&&  &  &  &  &  & \cellcolor{black} & \cellcolor{black} & \cellcolor{black} &  &  &  &  &  & \cellcolor{black} &  &  &  & \cellcolor{black} &  &   &\\
\hline
EQFM &   \cellcolor{black}&&  &  &  &  &  & \cellcolor{black} & \cellcolor{black} & \cellcolor{black} &  &  &  &  &  & \cellcolor{black} &  &  &  & \cellcolor{black} &  &   &\\
\hline
DLFM &   \cellcolor{black}&&  &  &  &  &  &  & \cellcolor{black} & \cellcolor{black} &  &  &  &  &  & \cellcolor{black} &  &  &  &  &  &   &\\
\hline
MLFM &   \cellcolor{black}&&  &  &  &  &  &  & \cellcolor{black} & \cellcolor{black} &  &  &  &  &  & \cellcolor{black} &  &  &  &  &  &   &\\
\hline
2FSK &   \cellcolor{black}& \cellcolor{black} &  &  &  &  &  & \cellcolor{black} & \cellcolor{black} & \cellcolor{black} &  &  & \cellcolor{black} & \cellcolor{black} &  & \cellcolor{black} &  &  &  & \cellcolor{black} &  &   &\\
\hline
4FSK &   \cellcolor{black}&&  &  &  &  &  & \cellcolor{black} & \cellcolor{black} & \cellcolor{black} &  &  &  &  &  & \cellcolor{black} &  &  &  &  &  &   &\\
\hline
COSTAS &  \cellcolor{black}&\cellcolor{black} & \cellcolor{black} &  & \cellcolor{black} & \cellcolor{black} & \cellcolor{black} &  &  &  & \cellcolor{black} & \cellcolor{black} &  &  & \cellcolor{black} &  & \cellcolor{black} & \cellcolor{black} & \cellcolor{black} &  & \cellcolor{black} & \cellcolor{black}  &\\
\hline 
SINFM & & &  & \cellcolor{black} &  &  &  &  &  &  &  &  &  &  &  &  &  &  &  &  &  &   &\\
\hline
TRIFM & & &  & \cellcolor{black} &  &  &  &  &  &  &  &  &  &  &  &  &  &  &  &  &  &   &\\
\hline
SCR &   &&  &  &  &  &  &  &  &  &  &  &  &  &  &  & \cellcolor{black} &  &  &  &  &   &\\
\hline
PCR3 &   &&  &  &  &  &  &  &  &  &  &  &  &  &  &  & \cellcolor{black} &  &  &  &  &   &\\
\hline
QFSK &   &&  &  &  &  &  &  &  &  &  &  & \cellcolor{black} &  &  &  &  &  &  &  &  &   &\\
\hline
\end{tabular}
}
\end{table*}

\begin{table*}[htbp]
\caption{Phase Modulation class usage across referenced studies. Each row represents a modulation type and each column a cited paper. Filled cells indicate the modulation was included in that study; empty cells indicate it was not.}
\label{tab:mod_presence_pm}
\centering
\footnotesize
\resizebox{\textwidth}{!}{%
\begin{tabular}{|l|l|c|c|c|c|c|c|c|c|c|c|c|c|c|c|c|c|c|c|c|c|c|l|}
\hline
 &  AIMC-Spec&\cite{RN161} & \cite{RN162} & \cite{RN152} & \cite{RN163} & \cite{RN151} & \cite{RN159} & \cite{RN160} & \cite{RN153} & \cite{RN149} & \cite{RN157} & \cite{RN156} & \cite{RN136} & \cite{RN148} & \cite{Hoang2019AutomaticRO} & \cite{Qu2018RadarSI} & \cite{Wang2017AutomaticRW} & \cite{Kong2018AutomaticLR} & \cite{Zhang2017ConvolutionalNN} & \cite{Tang2025AutomaticRO} & \cite{Chen2024LPIRS} & \cite{Luo2024RadarWR}  &\cite{Bhatti2024TransformerbasedMF}\\
\hline
BPSK &  \cellcolor{black}&\cellcolor{black} &  & \cellcolor{black} & \cellcolor{black} & \cellcolor{black} & \cellcolor{black} & \cellcolor{black} & \cellcolor{black} & \cellcolor{black} & \cellcolor{black} &  & \cellcolor{black} & \cellcolor{black} & \cellcolor{black} & \cellcolor{black} &  & \cellcolor{black} & \cellcolor{black} &  & \cellcolor{black} & \cellcolor{black}  &\\
\hline
QPSK &  \cellcolor{black}&\cellcolor{black} &  & \cellcolor{black} &  &  &  &  &  &  &  &  &  & \cellcolor{black} &  &  &  &  &  &  &  &   &\\
\hline
FRANK &  \cellcolor{black}&\cellcolor{black} & \cellcolor{black} &  & \cellcolor{black} & \cellcolor{black} & \cellcolor{black} & \cellcolor{black} & \cellcolor{black} & \cellcolor{black} & \cellcolor{black} & \cellcolor{black} &  &  & \cellcolor{black} & \cellcolor{black} &  & \cellcolor{black} & \cellcolor{black} & \cellcolor{black} & \cellcolor{black} & \cellcolor{black}  &\cellcolor{black}  \\
\hline
BARKER &   \cellcolor{black}&& \cellcolor{black} &  &  &  &  &  &  &  &  & \cellcolor{black} &  &  &  &  &  &  &  &  &  &   &\cellcolor{black}  \\
\hline
P1-4 &  \cellcolor{black}&\cellcolor{black} & \cellcolor{black} &  & \cellcolor{black} & \cellcolor{black} & \cellcolor{black} &  &  & \cellcolor{black} & \cellcolor{black} & \cellcolor{black} &  &  & \cellcolor{black} &  &  & \cellcolor{black} &  &  & \cellcolor{black} & \cellcolor{black}  &\cellcolor{black}  \\
\hline
T1-4 &   && \cellcolor{black} &  & \cellcolor{black} &  & \cellcolor{black} &  &  &  & \cellcolor{black} & \cellcolor{black} &  &  & \cellcolor{black} &  &  & \cellcolor{black} & \cellcolor{black} &  & \cellcolor{black} & \cellcolor{black}  &\\
\hline
\end{tabular}
}
\end{table*}

\begin{table*}[htbp]
\caption{Modulation class usage across referenced studies. Each row represents a modulation type and each column a cited paper. Filled cells indicate the modulation was included in that study; empty cells indicate it was not.}
\label{tab:mod_presence_hybrid}
\centering
\footnotesize
\resizebox{\textwidth}{!}{%
\begin{tabular}{|l|l|c|c|c|c|c|c|c|c|c|c|c|c|c|c|c|c|c|c|c|c|c|l|}
\hline
 &  AIMC-Spec&\cite{RN161} & \cite{RN162} & \cite{RN152} & \cite{RN163} & \cite{RN151} & \cite{RN159} & \cite{RN160} & \cite{RN153} & \cite{RN149} & \cite{RN157} & \cite{RN156} & \cite{RN136} & \cite{RN148} & \cite{Hoang2019AutomaticRO} & \cite{Qu2018RadarSI} & \cite{Wang2017AutomaticRW} & \cite{Kong2018AutomaticLR} & \cite{Zhang2017ConvolutionalNN} & \cite{Tang2025AutomaticRO} & \cite{Chen2024LPIRS} & \cite{Luo2024RadarWR}  &\cite{Bhatti2024TransformerbasedMF}\\
\hline
LFM-BPSK & & &  &  &  &  &  &  & \cellcolor{black} &  &  &  &  &  &  & \cellcolor{black} &  &  &  &  &  &   &\\
\hline
2FSK-BPSK & & &  &  &  &  &  &  & \cellcolor{black} &  &  &  &  &  &  & \cellcolor{black} &  &  &  &  &  &   &\\
\hline
LFM-SFM & & &  &  &  &  &  &  &  & \cellcolor{black} &  &  &  &  &  &  &  &  &  &  &  &   &\\
\hline
BPSK-LFM &   &&  &  &  &  &  &  &  &  &  &  &  &  &  &  &  &  &  & \cellcolor{black} & \cellcolor{black} &   &\\
\hline
EQFM-BPSK &   &&  &  &  &  &  &  &  &  &  &  &  &  &  &  &  &  &  & \cellcolor{black} &  &   &\\
\hline
EQFM-2FSK &   &&  &  &  &  &  &  &  &  &  &  &  &  &  &  &  &  &  & \cellcolor{black} &  &   &\\
\hline
FRANK-EQFM &   &&  &  &  &  &  &  &  &  &  &  &  &  &  &  &  &  &  & \cellcolor{black} &  &   &\\
\hline
FRANK-4FSK &   &&  &  &  &  &  &  &  &  &  &  &  &  &  &  &  &  &  & \cellcolor{black} &  &   &\\
\hline
LFM-4FSK &   &&  &  &  &  &  &  &  &  &  &  &  &  &  &  &  &  &  & \cellcolor{black} &  &   &\\
\hline
BPSK-FSK &   &&  &  &  &  &  &  &  &  &  &  &  &  &  &  &  &  &  &  & \cellcolor{black} &   &\\
\hline
FSK-LFM &   &&  &  &  &  &  &  &  &  &  &  &  &  &  &  &  &  &  &  &  & \cellcolor{black}  &\\
\hline
FSK-BPSK &   &&  &  &  &  &  &  &  &  &  &  &  &  &  &  &  &  &  &  &  & \cellcolor{black}  &\\
\hline
\end{tabular}
}
\end{table*}

\subsection{Signal to Noise Ratio (SNR)}
As the dataset is synthetically generated, it lacks natural noise. To approximate real-world conditions, Additive White Gaussian Noise (AWGN) is applied dynamically during data loading through a custom Python dataset class, rather than being permanently stored with the generated signals. AWGN was selected as it models thermal noise, which is commonly encountered in communication and radar systems due to its random nature and uniform spectral distribution. The dataset includes five SNR levels ranging from +6 to –6\,dB in steps of 3\,dB, specifically: +6, +3, 0, –3, and –6\,dB.

Noise is added directly to the raw complex I/Q pulse prior to spectrogram transformation. Unlike the previous implementation, the SNR calculation is based solely on the average power of the pulse itself, excluding any zero-padding or non-signal regions. This ensures that the specified SNR accurately reflects the true signal energy content. An SNR of 0\,dB therefore represents equality between the average pulse power and the noise power. Figure~\ref{fig:snr-samples} presents representative spectrograms across varying SNR levels, illustrating the progressive degradation in visual clarity as noise increases. While AIMC-Spec currently models only AWGN, future releases will incorporate additional channel effects—such as multipath propagation, fading, and pulse jitter—to more closely emulate operational radar environments.

\subsection{Intrapulse Modulations}
AIMC-Spec includes 30 intrapulse modulation types grouped into frequency (15) and phase (15) categories. These modulation types also serve as the class labels for classification tasks. FM signals are often visually distinct in spectrograms due to frequency sweeps. In contrast, PM types produce more subtle patterns—phase shifts introduce discontinuities or faint texture shifts that are less salient to image-based classifiers.

\subsection{Data Set Structure}
The dataset is distributed as two directories: a training directory and a testing directory. Each directory contains 30 binary \texttt{.h5} files (one per intrapulse modulation), stored using Python’s \texttt{h5py} library. Each \texttt{.h5} file contains a collection of signals, where each entry comprises a single pulse represented by complex I/Q samples together with associated metadata. The total size of the training directory is 28\,GB, while the testing directory is 288\,MB.

The dataset was originally intended to support a randomly generated training split alongside a fixed, pre-defined test split. However, due to constraints associated with releasing the full dataset generation pipeline, this approach was not feasible. To address this, the released training data were generated at a substantially larger scale than required for a standard 80:20 split. A deterministic sub-sampling procedure is therefore provided to construct user-defined train/validation partitions at run time. Specifically, for a given random seed and target ratio (e.g., 80:20), the loader samples disjoint subsets from each modulation file without replacement, yielding a reproducible split while preserving per-modulation balance.

\section{Benchmark Algorithms}

\begin{figure}[h!]
    \centering
    \includegraphics[width=0.9\linewidth]{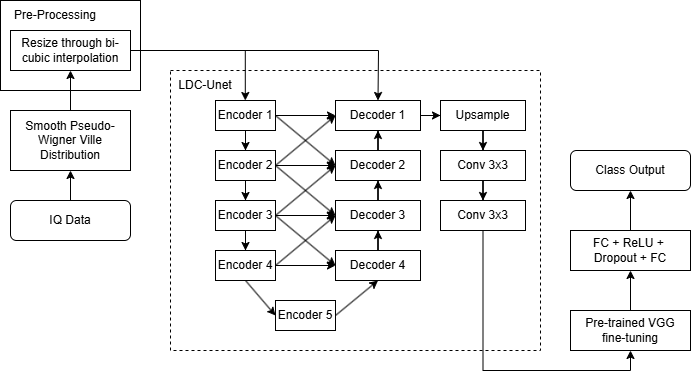}
    \caption{Compressed architecture of LDC-Unet.}
    \label{fig:ldc-unet-arch}
\end{figure}

\begin{figure}[h!]
    \centering
    \includegraphics[width=0.9\linewidth]{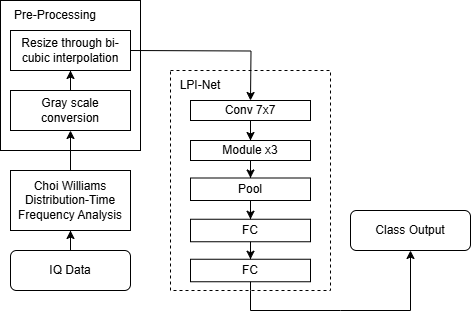}
    \caption{Compressed architecture of LPI-Net.}
    \label{fig:lpi-net-arch}
\end{figure}

\begin{figure}[h!]
    \centering
    \includegraphics[width=0.9\linewidth]{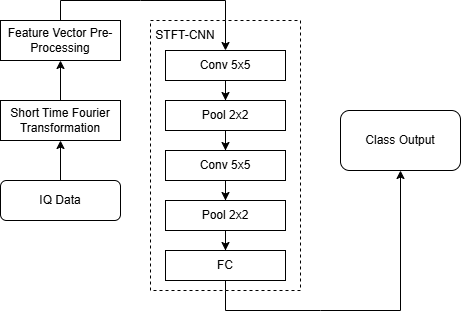}
    \caption{Compressed architecture of STFT-CNN.}
    \label{fig:stft-cnn-arch}
\end{figure}

A curated selection of five algorithms was chosen to benchmark AIMC-Spec, reflecting both their influence in the field and their architectural diversity. The set spans convolutional, denoising, U-Net-like, pretrained, and transformer architectures, ensuring that different feature-extraction paradigms are represented. Each has demonstrated strong performance in prior studies under varying SNR conditions. Collectively, these models form a balanced cross-section of contemporary AIMC research and provide insight into how distinct neural architectures handle a common spectrogram representation. Figure~\ref{fig:preprocessing-images} shows the preprocessing outputs for each model on a 4FSK signal at 0\,dB SNR, with architectural summaries provided in Appendix~\ref{app:architectures} (Figures~\ref{fig:cdae-arch}–\ref{fig:stft-cnn-arch}).

\subsection{LDC-Unet: Unet + VGG19}
The method in \cite{RN163} employs the Smooth Pseudo-Wigner Ville Distribution (SPWVD) to generate time-frequency images, which are resized to 128\,$\times$\,128 using bicubic interpolation—an approach that averages pixel values across a neighborhood for smooth visual scaling. The model combines a modified U-Net with local dense connections and a fine-tuned VGG-based deep CNN. It is trained using a custom SSR Loss and optimized via stochastic gradient descent (SGD). An example of the resulting spectrogram is shown in Figure~\ref{fig:ldc_unet}.

\subsection{LPI-Net: Modular Lightweight CNN}
The architecture in \cite{RN162} uses the CWD for spectrogram generation. As with LDC-Unet, outputs are resized to 64\,$\times$\,64 using bicubic interpolation and converted to grayscale. The model is a modular CNN composed of repeated convolutional blocks, trained using cross-entropy loss and optimized with SGD. Its grayscale pipeline and modular design contribute to its computational efficiency. The preprocessed 4FSK example is displayed in Figure~\ref{fig:lpi_net}.

\subsection{CDAE-DCNN: Denoising Auto-Encoder + DCNN}
In \cite{RN153}, spectrograms are produced using the CTFD, resized to 64\,$\times$\,64 via linear interpolation, and normalized. A convolutional denoising autoencoder first removes noise from the spectrograms, which are then passed to a deeper CNN for classification. The model uses the Adam optimizer with standard cross-entropy loss. The denoised spectrogram output is shown in Figure~\ref{fig:cdae_dcnn}.

\subsection{STFT-CNN: Simple CNN}
The approach in \cite{RN136} uses the simplest architecture in the benchmark set, comprising only 5 layers — the fewest among all models compared. It generates spectrograms via the STFT, resizes them using pixel area interpolation, and applies normalization. An iterative noise suppression step is used before classification by a lightweight CNN trained with Adam and cross-entropy loss. The corresponding spectrogram image is presented in Figure~\ref{fig:stft_cnn}.

\subsection{ViT: Phase-Spectrum Vision Transformer}
\begin{figure*}[t]
\centering
\subfigure[Raw 4FSK]{\includegraphics[width=0.30\textwidth]{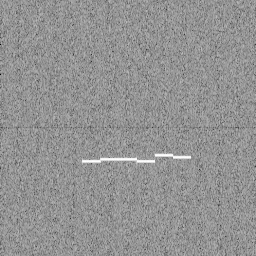}\label{fig:raw_4fsk}}
\hfill
\subfigure[LDC-Unet]{\includegraphics[width=0.30\textwidth]{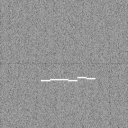}\label{fig:ldc_unet}}
\hfill
\subfigure[LPI-Net]{\includegraphics[width=0.30\textwidth]{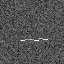}\label{fig:lpi_net}}
\\[2pt]
\subfigure[CDAE-DCNN]{\includegraphics[width=0.30\textwidth]{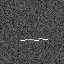}\label{fig:cdae_dcnn}}
\hfill
\subfigure[STFT-CNN]{\includegraphics[width=0.30\textwidth]{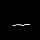}\label{fig:stft_cnn}}
\hfill
\subfigure[ViT]{%
  \raisebox{0.25\height}{\includegraphics[width=0.30\textwidth]{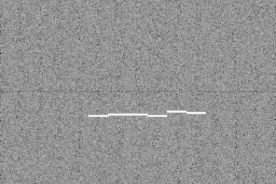}}%
  \label{fig:vit}
}

\caption{Spectrogram representations of the 4FSK modulation at 0\,dB SNR following preprocessing by the six benchmark algorithms. Each spectrogram was generated using a 256-point Hann window with 50\,\% overlap.}
\label{fig:preprocessing-images}
\vspace{-3mm} 
\end{figure*}

The ViT model from \cite{Bhatti2024TransformerbasedMF} operates on phase spectrograms extracted via the STFT, focusing on the carrier-aligned phase row (typically the seventh) to form compact one-dimensional images that capture phase variations over time. Because only this single frequency row corresponding to the carrier is retained, the resulting phase-spectrum image is inherently non-square rather than the conventional \(N\times N\) time–frequency map. This compact 1D representation significantly reduces computational complexity while preserving modulation-specific phase patterns. These cropped phase spectra are divided into 23-sample patches and passed through eight transformer layers with four attention heads. A learnable classification token and multilayer perceptron head perform the final prediction. Trained using cross-entropy loss and the AdamW optimizer, the model achieves over 92\% accuracy at –16\,dB on six phase-coded modulations. A sample non-square input is shown in Figure~\ref{fig:vit}.

Together, these algorithms represent a diverse set of modeling strategies, ranging from lightweight CNNs to transformers, covering the breadth of contemporary AIMC techniques.

\section{Experimenting and Benchmarking}
\subsection{Methodology}

This study benchmarks five representative AIMC algorithms using a unified FFT-based spectrogram pipeline implemented with open-source signal-processing tools. Each model was trained and tested using the same spectrogram input type to ensure consistency and isolate architectural performance. Several prior works lacked sufficient methodological detail or were designed for markedly different data conditions (e.g. short I/Q segments rather than full-scale pulse windows). In addition, attempts to directly apply those transforms to the AIMC-Spec signals produced distorted or invalid time-frequency images. Using an FFT-based approach with 50\% window overlap provides stable, reproducible inputs and practical runtime. 

While other time-frequency transforms such as CWD, SPWVD, and FSST can yield richer-time frequency resolution, these were intentionally excluded to prevent confounding architectural comparison with transform-specific effects. Standardizing on the FFT transform isolates model behavior from representation specific tuning, aligning with a primary objective of this work--to examine how distinct neural architectures (CNN, UNet, ViT, etc.) perform on the AIMC-Spec dataset under identical input conditions. A follow-up study will explore how alternative transforms influence model performance on AIMC-Spec. 

\subsection{Implementation}

All spectrograms were generated using the open-source \texttt{cupy} Python library to accelerate FFT computation on GPU. During early experimentation, on-the-fly spectrogram generation was identified as a significant computational bottleneck, substantially increasing training time. To mitigate this, spectrograms were precomputed offline and stored on disk prior to model training. Memory-mapped (memmap) files were then used to efficiently stream spectrogram data during training, enabling scalable access without loading the full dataset into RAM. This approach significantly improved I/O efficiency and overall training throughput.

A consistent 80:20 train–test split was applied to each model. Original loss functions and optimizers were preserved as described in their respective source papers.

Training was performed on an NVIDIA GeForce RTX 4090 GPU using a batch size of 32. Each model trained for up to 50 epochs, with early stopping (patience of 10 epochs) to prevent overfitting. During training, loss histories were recorded for each run. Final model evaluations included overall classification accuracy and confusion matrix generation.

\subsection{Experiments}

Experiments were designed to evaluate model robustness under varying noise conditions. Each model was trained on a combined dataset spanning five SNR levels (+6, +3, 0, –3, and –6\,dB) and subsequently evaluated at each individual SNR level. This setup assesses the model’s ability to generalize across different noise intensities while maintaining performance under specific operating conditions.

Each experimental configuration was conducted in two variants. The first included all 30 intrapulse modulation types within AIMC-Spec, consisting of 15 frequency-modulated and 15 phase-modulated signals, to evaluate performance across the complete classification task. The second restricted the task to the subset of 15 frequency-modulated signals, selected for their well-defined spectral characteristics and higher separability in spectrogram representations.

Together, these two variants provide both a comprehensive benchmark across the full modulation space and a focused assessment under more controlled classification conditions.

\begin{figure}[bt]
    \centering
    \begin{minipage}{\linewidth}
        \raggedleft
        \includegraphics[width=\linewidth]{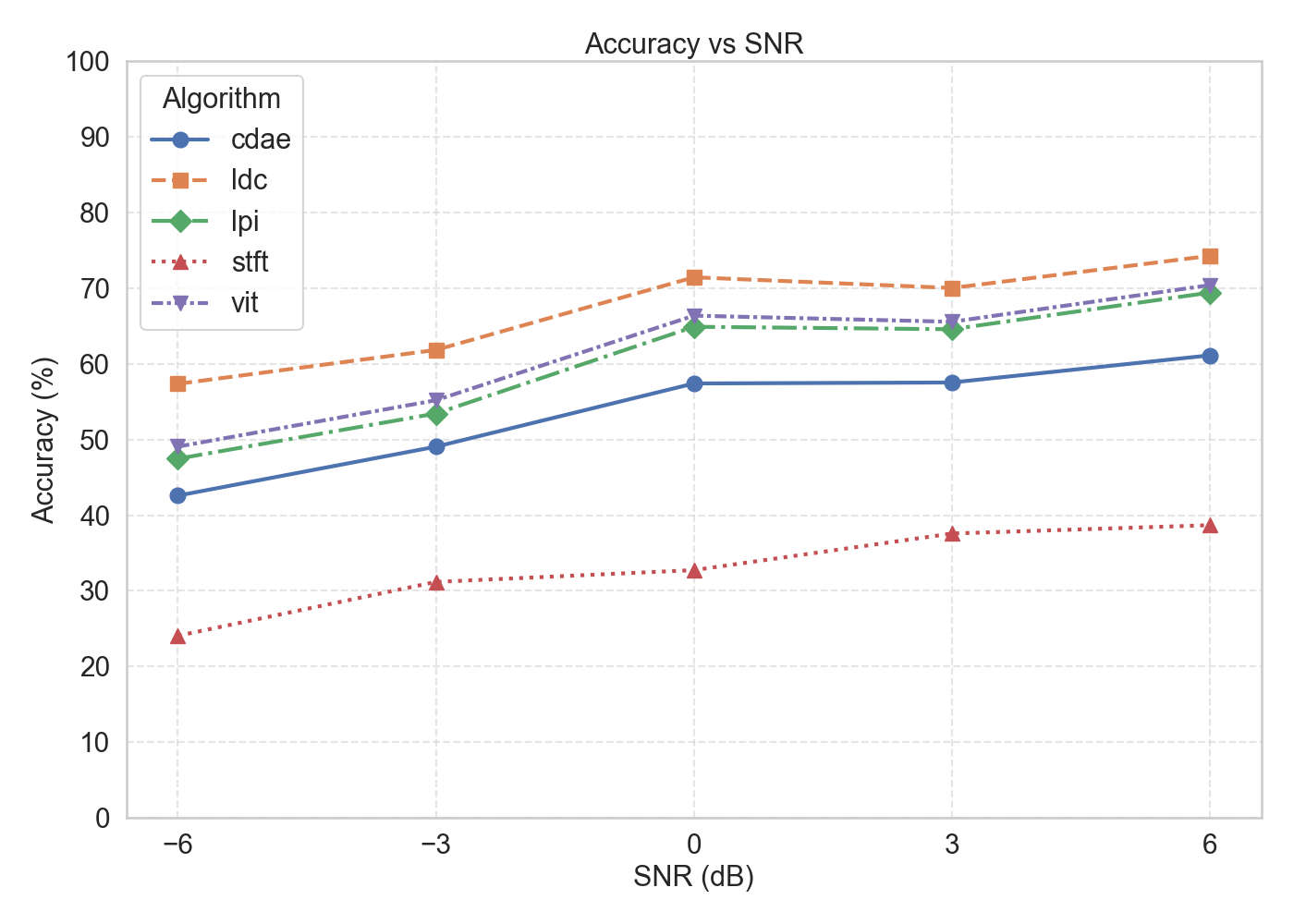}
        \caption{Accuracy (\%) vs SNR (dB) for experiment SNR($x$):SNR($x$), using all modulation types.}
        \label{fig:results-all}
        
        \vspace{0.6em}

        \includegraphics[width=\linewidth]{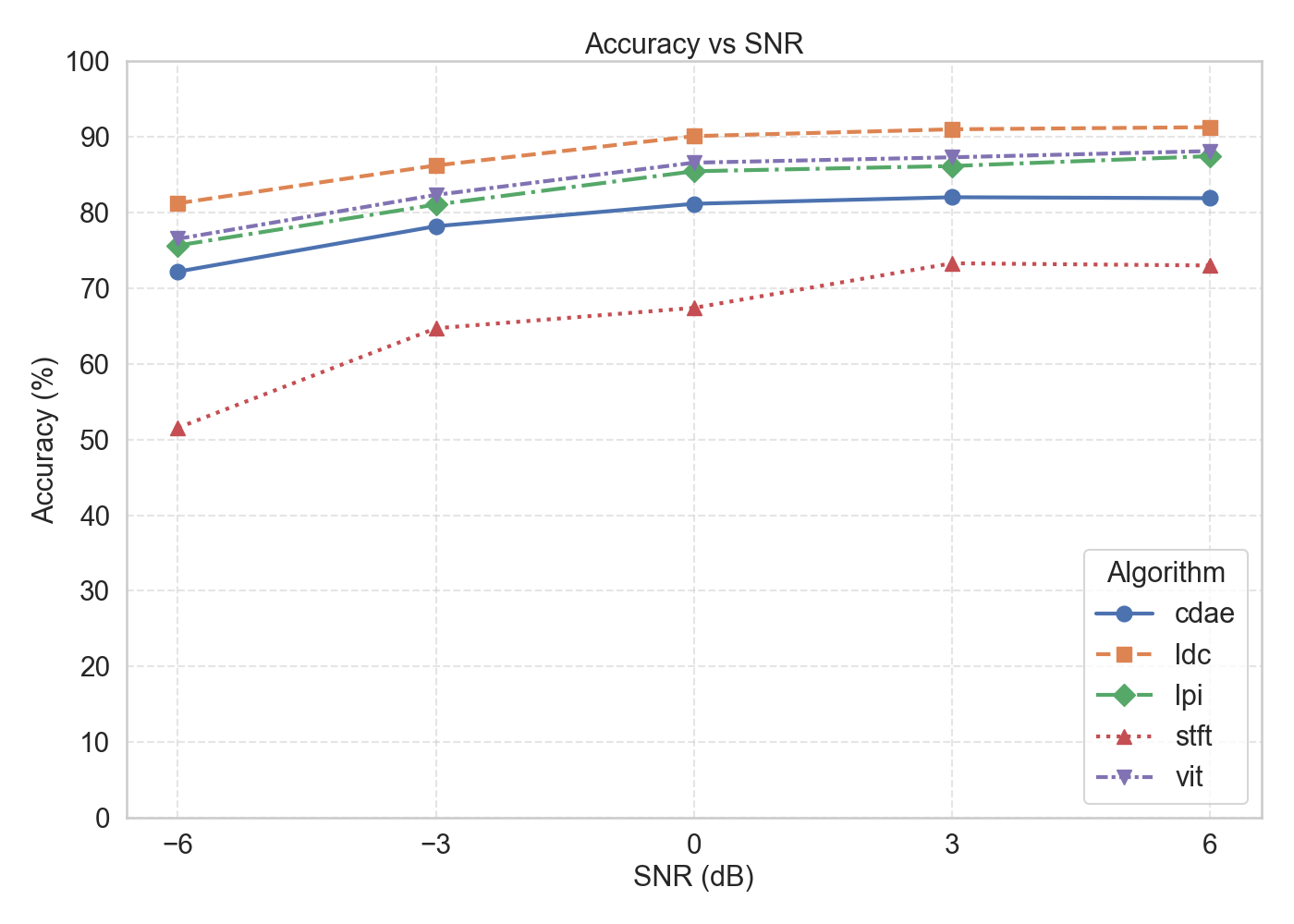}
        \caption{Accuracy (\%) vs SNR (dB) for experiment SNR(ALL):SNR($x$), using all modulation types.}
        \label{fig:results-fm}
    \end{minipage}
\end{figure}

\subsection{Results}

\subsubsection*{Algorithm Performance Across SNR Levels}

LDC-Unet achieved the strongest performance across all evaluated conditions for both the full modulation classification task and the FM subset, as shown in Figures~\ref{fig:results-all} and~\ref{fig:results-fm}. On the full modulation set, it obtained a mean accuracy of 66.91\%, with performance increasing from 57.37\% at –6\,dB to 74.29\% at +6\,dB. On the FM subset, the model achieved a mean accuracy of 87.95\%, ranging from 81.24\% at –6\,dB to 91.30\% at +6\,dB. Across all SNR levels and both experimental tasks, LDC-Unet consistently maintained the highest accuracy.

The next group of models consisted of ViT, LPI-Net, and CDAE-DCNN, which followed the same overall ranking in both experiments. ViT achieved the second-best results, reaching a mean accuracy of 61.19\% on the full modulation task and 84.07\% on the FM subset, followed closely by LPI-Net with 59.72\% and 83.05\%, respectively. CDAE-DCNN ranked fourth overall, achieving 53.39\% mean accuracy on the full task and 78.95\% on the FM subset. Although these models exhibited steady improvements as SNR increased, their performance remained consistently below that of LDC-Unet across the evaluated noise levels.

The STFT-CNN baseline produced the lowest performance across both experiments. It achieved a mean accuracy of 32.62\% on the full modulation task, ranging from 24.06\% at –6\,dB to 38.68\% at +6\,dB, and 65.80\% on the FM subset, increasing from 51.56\% to 73.01\% across the same SNR range. Despite this improvement with increasing SNR, STFT-CNN remained substantially below the other models. Across all algorithms, accuracy increased steadily with SNR, and the relative ranking of models remained consistent, with LDC-Unet outperforming ViT, LPI-Net, CDAE-DCNN, and STFT-CNN throughout the evaluated range.

\subsubsection*{Algorithm Trends}

LDC-Unet consistently achieved the highest accuracy across all evaluated SNR levels in both experimental tasks. Among the remaining models, ViT reported the second-highest overall accuracy, followed closely by LPI-Net and CDAE-DCNN, while STFT-CNN produced the lowest performance across the evaluated conditions. Despite its comparatively compact architecture, STFT-CNN still demonstrated that meaningful classification performance can be achieved with lightweight CNN models when paired with structured time–frequency representations.

Across all algorithms, accuracy increased steadily as SNR improved, with the most pronounced gains occurring between –6\,dB and 0\,dB. The relative ranking of models remained consistent across the evaluated noise levels, with LDC-Unet outperforming ViT, LPI-Net, CDAE-DCNN, and STFT-CNN throughout the entire range. Despite these improvements at higher SNRs, the overall accuracy levels reflect the challenging nature of the AIMC-Spec dataset, which includes a large number of modulation types with overlapping spectral characteristics and realistic noise conditions. This complexity introduces significant inter-class ambiguity, particularly for phase-modulated signals, making robust classification under low-SNR conditions substantially more difficult than in many previously reported AIMC benchmarks.

\subsubsection*{Analytical Discussion}

While overall accuracies provide a useful benchmark, deeper inspection of the accuracy–SNR trends in Figures~\ref{fig:results-all} and~\ref{fig:results-fm} reveals distinct behavioral patterns that align with each algorithm’s design.

LDC-Unet maintained the highest stability across all noise conditions, with performance remaining relatively consistent across the evaluated SNR range from –6\,dB to +6\,dB. This resilience stems from its local dense skip connections and residual encoder blocks, which preserve multi-scale features and reconstruct fine-grained spectrogram detail even when noise dominates the input. The use of SSR-loss further enforces intra-class compactness and inter-class separation, allowing the model to maintain discriminative boundaries under low-SNR degradation.

CDAE-DCNN also performed strongly at lower SNRs due to its denoising pre-training stage. However, compared with LDC-Unet, it exhibited a steeper performance decline as noise increased. This behavior is consistent with the CDAE’s tendency to over-smooth high-frequency or phase-rich modulations, which can reduce discriminative detail in complex spectrogram structures.

ViT achieved competitive performance but remained slightly below the CNN-based architectures across both experimental tasks. Transformer models rely heavily on large and diverse training datasets to learn robust global attention patterns. While this mechanism can capture long-range dependencies in spectrogram images, it is also more sensitive to noise-induced feature variability when training data diversity is limited.

STFT-CNN and LPI-Net demonstrated consistent but more modest performance, primarily limited by their input representations. STFT-CNN’s compact spectrogram images constrain the capture of long-term spectral structures, while LPI-Net’s grayscale preprocessing reduces feature richness for overlapping or phase-coded signals. Nevertheless, both models highlight that well-structured spectrogram inputs can still achieve reasonable classification performance using comparatively lightweight CNN architectures.

Across all algorithms, the FM-only task consistently produced higher accuracy than the full modulation classification task. This difference indicates that the principal challenge lies in recognizing phase and hybrid modulations, where spectral cues are less visually separable and inter-class boundaries overlap under noise. The different slopes of each model’s accuracy–SNR curve further highlight architectural sensitivity to these factors: models with skip connections or denoising mechanisms degrade more gradually, whereas architectures with more constrained feature extraction exhibit sharper performance declines as noise increases.

Collectively, these observations suggest that model behavior on AIMC-Spec is driven less by absolute accuracy and more by how effectively each architecture preserves discriminative information under noise. This insight is particularly relevant for guiding future multi-frame and multi-modal model designs, where temporal and cross-representation cues may further improve robustness.

\subsubsection*{Influence of Preprocessing and Input Design}

The preprocessing choices and input formats—previously detailed for each algorithm—appeared to influence classification outcomes. Algorithms that preserved more visual detail through color channels or higher-resolution spectrogram representations tended to achieve higher accuracy on more complex modulation types. For example, LDC-Unet and CDAE-DCNN were better able to retain phase and hybrid modulation characteristics, likely due to their use of RGB inputs and larger spectrogram resolutions.

By contrast, LPI-Net’s grayscale preprocessing likely reduced feature richness, particularly for modulations with overlapping frequency or phase behaviors. STFT-CNN, although using RGB inputs, was constrained by its small image size, limiting the spatial representation of spectral features. ViT, which received phase spectrogram inputs in a single-channel format, achieved slightly lower overall accuracy compared with the leading CNN-based models, suggesting that the quantity and richness of input information may play an important role in transformer-based radar classification tasks.

These observations suggest that aligning input representation with architectural characteristics may improve robustness, particularly in noisy environments where subtle spectral cues become more difficult to distinguish.

\section{Conclusion}

This paper presents AIMC-Spec, a synthetic dataset designed to support benchmarking of AIMC using spectrogram-based image classification techniques. The dataset includes 30 intrapulse modulation types spanning frequency and phase categories, and evaluates performance across a controlled SNR range from –6\,dB to +6\,dB. By combining diverse modulation classes with realistic noise conditions, AIMC-Spec provides a structured environment for assessing model robustness under varying signal quality.

Five representative deep learning algorithms were implemented and evaluated on AIMC-Spec, including LDC-Unet, ViT, LPI-Net, CDAE-DCNN, and STFT-CNN. The benchmarking experiments revealed clear differences in performance across model architectures, with LDC-Unet consistently achieving the highest accuracy across both the full modulation task and the FM subset. CNN-based architectures with dense connections and multi-scale feature extraction demonstrated greater resilience to noise, while transformer and lightweight CNN approaches showed comparatively reduced robustness.

Overall, the experimental results highlight the complexity of the AIMC-Spec classification task, particularly when distinguishing phase and hybrid modulations whose spectral characteristics overlap under noisy conditions. These findings demonstrate that AIMC-Spec can serve not only as a benchmarking dataset but also as a diagnostic tool for understanding how different model architectures respond to challenging signal environments.

\subsection{Future Works}

Future work will extend AIMC-Spec by incorporating additional intrapulse modulation types and expanding the diversity of signal conditions used during dataset generation. Including a broader range of modulation behaviors will allow more comprehensive benchmarking and help evaluate model performance across increasingly complex classification tasks.

Beyond architectural benchmarking, future iterations of AIMC-Spec will incorporate greater environmental realism by integrating measured or parameter-derived radar signals and by simulating operational effects such as multipath propagation, fading, pulse dropouts, phase jitter, and measurement noise. These additions will enable the dataset to emulate real-world conditions more accurately and provide a stronger basis for evaluating algorithm robustness and generalization performance.

With continued development, AIMC-Spec can evolve into a comprehensive benchmark for the AIMC community. Its structured and transparent design supports reproducible experimentation, fair comparison between algorithms, and deeper analysis of how model architectures respond to challenging signal environments.


\bibliographystyle{IEEEtran}
\bibliography{references.bib}

\appendices
\section{Compressed Benchmark Architectures}
\label{app:architectures}

\begin{figure}[h!]
    \centering
    \includegraphics[width=0.9\linewidth]{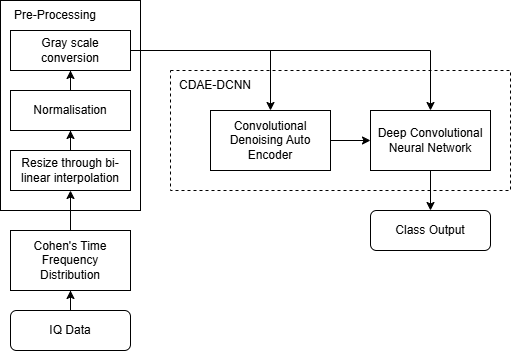}
    \caption{Compressed architecture of CDAE-DCNN.}
    \label{fig:cdae-arch}
\end{figure}

\begin{figure}[h!]
    \centering
    \includegraphics[width=0.9\linewidth]{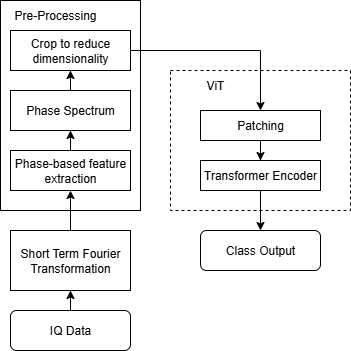}
    \caption{Compressed architecture of ViT.}
    \label{fig:vit-arch}
\end{figure}

\vfill\break
\section{Tabular Format of Graphs}

\begin{table}[H]
\caption{Accuracy (\%) vs SNR (dB) for the full modulation classification task. Best values per row are in bold.}
\label{tab:all_accuracy}
\centering
\resizebox{\columnwidth}{!}{%
\begin{tabular}{|c|c|c|c|c|c|}
\hline
\textbf{SNR} & \textbf{STFT-CNN} & \textbf{LPI-Net} & \textbf{CDAE-DCNN} & \textbf{LDC-Unet} & \textbf{ViT} \\
\hline
$-6$ & 24.06 & 47.47 & 42.59 & \textbf{57.37} & 49.07 \\
\hline
$-3$ & 31.18 & 53.44 & 49.06 & \textbf{61.85} & 55.19 \\
\hline
$0$  & 32.74 & 64.91 & 57.42 & \textbf{71.45} & 66.38 \\
\hline
$+3$ & 37.58 & 64.59 & 57.55 & \textbf{70.04} & 65.57 \\
\hline
$+6$ & 38.68 & 69.43 & 61.13 & \textbf{74.29} & 70.42 \\
\hline
\end{tabular}%
}
\end{table}

\begin{table}[H]
\caption{Accuracy (\%) vs SNR (dB) for the frequency-modulation-only classification task. Best values per row are in bold.}
\label{tab:fm_accuracy}
\centering
\resizebox{\columnwidth}{!}{%
\begin{tabular}{|c|c|c|c|c|c|}
\hline
\textbf{SNR} & \textbf{STFT-CNN} & \textbf{LPI-Net} & \textbf{CDAE-DCNN} & \textbf{LDC-Unet} & \textbf{ViT} \\
\hline
$-6$ & 51.56 & 75.67 & 72.21 & \textbf{81.24} & 76.54 \\
\hline
$-3$ & 64.71 & 81.07 & 78.21 & \textbf{86.24} & 82.35 \\
\hline
$0$  & 67.41 & 85.47 & 81.17 & \textbf{90.12} & 86.61 \\
\hline
$+3$ & 73.31 & 86.17 & 82.04 & \textbf{91.02} & 87.32 \\
\hline
$+6$ & 73.01 & 87.47 & 81.91 & \textbf{91.30} & 88.14 \\
\hline
\end{tabular}%
}
\end{table}

\newpage

\clearpage
\section{List of Acronyms}

\begin{table}[ht]
\centering
\caption{Intrapulse Modulation Types and Descriptions}
\resizebox{\textwidth}{!}{%
\begin{tabular}{|c|c|c|p{7cm}|}
\hline
\textbf{Acronym} & \textbf{AIMC-Spec} & \textbf{Full Name} & \textbf{Description} \\
\hline

UNMOD & \ding{51} & Unmodulated & Constant frequency, no modulation across the pulse duration. \\
\hline
LFM & \ding{51} & Linear Frequency Modulation & Frequency increases (LFM\_up) or decreases (LFM\_down) linearly across time (chirp). \\
\hline
NLFM & \ding{51} & Nonlinear Frequency Modulation & Frequency changes non-linearly to reduce sidelobes. \\
\hline
SFM & \ding{51} & Step Frequency Modulation & Discrete frequency steps within the pulse. \\
\hline
EQFM & \ding{51} & Equal-Width Frequency Modulation & Uniform sub-bands of frequency modulation. \\
\hline
DLFM & \ding{51} & Discrete Linear FM & Linear FM in discrete blocks, can be either up\_down or down\_up. \\
\hline
MLFM & \ding{51} & Modified Linear FM & Linear frequency modulation with nonlinear terms added to reduce sidelobes and enhance range resolution.
 \\
\hline
BFSK & \ding{51} & Binary Frequency Shift Keying & Equivalent to 2FSK; binary frequency hopping. \\
\hline
4FSK & \ding{51} & 4-level Frequency Shift Keying & Switches between four frequencies. \\
\hline
FSK & \ding{55} & Frequency Shift Keying & General category of modulation using frequency switching. \\
\hline
SINFM & \ding{55} & Sinusoidal Frequency Modulation & Frequency follows a sinusoidal curve across time. \\
\hline
TRIFM & \ding{55} & Triangular Frequency Modulation & Symmetric linear up/down chirp within a pulse. AIMC-Spec considers DLFM\_Up\_Down as TriFM due to the single pulse classification task. \\
\hline
SCR & \ding{55} & Stepped Chirp & Sequence of chirps with stepped frequency offsets. \\
\hline
PCR3 & \ding{55} & Piecewise Constant Ramp (3 segments) & Frequency ramps in fixed intervals across 3 segments. \\
\hline
QFSK & \ding{55} & Quadrature Frequency Shift Keying & Four FSK tones placed orthogonally. \\
\hline

BPSK & \ding{51} & Binary Phase Shift Keying & Phase alternates between two states (0, $\pi$). \\
\hline
QPSK & \ding{51} & Quadrature Phase Shift Keying & Phase modulated using four discrete values. \\
\hline
FRANK & \ding{51} & Frank Code & Polyphase code with linear phase progression. \\
\hline
BARKER & \ding{51} & Barker Code & Binary code with good autocorrelation properties. Consists of: 2\_1, 2\_2, 3, 4\_1, 4\_2, 5, 7, 11, 13 \\
\hline
COSTAS & \ding{51} & Costas Code & Frequency-hopping sequence with good ambiguity properties. \\
\hline
P1–4 & \ding{51} & P1–P4 Polyphase Codes & Four distinct polyphase modulation schemes designed for pulse compression, each offering specific autocorrelation properties and sidelobe control for improved range resolution. \\
\hline
EXP & \ding{51} & Exponential FM & Frequency varies exponentially over time. \\
\hline
T1-4 & \ding{55} & T1–T4 Codes & Polyphase codes with tailored phase sequences for low sidelobes and Doppler tolerance, typically 16–100 chips long, similar to P1–P4 codes but optimized for Doppler performance.\\
\hline

LFM-BPSK & \ding{55} & LFM with BPSK & LFM chirp modulated with BPSK phase switching. \\
\hline
2FSK-BPSK & \ding{55} & 2FSK with BPSK & 2FSK combined with binary phase modulation. \\
\hline
LFM-SFM & \ding{55} & LFM with SFM & Hybrid of linear chirp and step frequency. \\
\hline
BPSK-LFM & \ding{55} & BPSK with LFM & BPSK applied over an LFM waveform. \\
\hline
EQFM-BPSK & \ding{55} & EQFM with BPSK & Phase modulated EQFM waveform. \\
\hline
EQFM-2FSK & \ding{55} & EQFM with 2FSK & Frequency hopping within EQFM envelope. \\
\hline
FRANK-EQFM & \ding{55} & Frank-coded EQFM & EQFM waveform modulated using Frank coding. \\
\hline
FRANK-4FSK & \ding{55} & Frank-coded 4FSK & 4FSK with Frank phase coding. \\
\hline
LFM-4FSK & \ding{55} & LFM and 4FSK hybrid & Combines LFM’s linear chirp with 4FSK’s four discrete frequency steps to improve range resolution and interference rejection. \\
\hline
BPSK-FSK & \ding{55} & BPSK with FSK & Binary phase shift applied to FSK waveform. \\
\hline
FSK-LFM & \ding{55} & FSK with LFM & Frequency-hopping waveform with LFM characteristics. \\
\hline
FSK-BPSK & \ding{55} & FSK with BPSK & Frequency shift keyed waveform with binary phase. \\
\hline
\end{tabular}
}
\label{tab:intrapulse-modulations}
\end{table}

\end{document}